# Aristotle's Original Idea
## For and Against Logic in the era of AI[1]

Antonis Kakas[2], University of Cyprus

**Summary**

Aristotle[3] is generally accepted as the father of logic. The ideas that he raised in his study of logical reasoning carried the development of science over the centuries. Any scientific theory's mathematical formalization is one that falls under his idea of Demonstrative Science. Today, in the era of AI, this title of the fatherhood of logic has a renewed significance. Behind it lies his original idea that human reasoning could be studied as a process and that perhaps there exist universal systems of reasoning that underly all human reasoning irrespective of the content of what we are reasoning about. This is a daring idea as it essentially says that the human mind can study itself and indeed that it has the capacity to unravel its own self. Irrespective of whether this is possible or not, it is a thought that is a prerequisite for the existence and development of Artificial Intelligence.

In this article, we look into Aristotle's work on human thought, his work on reasoning itself but also on how it relates to science and human endeavour more generally, from a modern perspective of Artificial Intelligence and ask if this can help enlighten our understanding of AI and Science more generally. The main observations drawn from this journey of his work are the following:

---

[1] This article is concerned with the large picture of Aristotle's work trying to view it within a modern perspective of AI and science. It does not aim to present the exact detailed form of his work, but rather to try to capture the essence of the main ideas relating to AI and how these could inform its future development. The article is addressed to those who have a general interest in philosophy and perhaps also to Aristotelian scholars who would be interested in his work from a modern perspective of AI. It is also hoped that AI researchers and practitioners will find it interesting to read when contemplating the larger scientific and social issues of AI.

[2] Co-founder of Argument Theory (https://www.argument-theory.com/), Paris, France.

[3] It is said that there are two Aristotles. One is the actual person that lived and worked in ancient times. The other is the person that we envisage as we study his work and try to understand how exactly he might have thought about matters: we ascribe to him an intellectual persona based on our interpretation of his work. A person that depends more on the particular perspective that we take on his work at a given time in history and the prevailing philosophical and societal cultures of the time. For non-Aristotelian scholars, as myself, when we write about Aristotle's work, we fall more into this second person category and therefore we are ready to stand corrected for any of the specifics of our presentation and reference to his work.

With regards to Aristotle, the main aim of this article is to give such a second person perspective of Aristotle from the viewpoint of Artificial Intelligence and to argue for a fresh new reading of Aristotle under this lens of AI. Our perspective will only be a partial one, concentrating more on the study of understanding human thinking and its relation to science and a limited view on the role of thinking on wider societal issues, such as that of ethics.

- Aristotle's original thought that human reasoning can be systematized can be understood as the embryonic start of AI.

- The dream of reason, as set by Aristotle, was in effect to rationalize reasoning as found widely in human thought and to use this in rational human behavior. The dream of AI is much like this dream of reason of Aristotle.

- Aristotle's notion of demonstrative necessity of logical reasoning with absolute certainty, that has been the bedrock of theoretical science till today, is not appropriate for AI.

- Despite the emphasis that he places on demonstrative necessity in reasoning he is also aware that a form of defeasible reasoning is needed within many fields. His studies of the role of human reasoning in our individual and societal endeavours reveal this need.

- A modern AI perspective on defeasible reasoning indicate that Aristotle's ideas on defeasible reasoning could be formalized through his notion of Dialectic Argument.

- Aristotle's Dialectic Argument has a universal nature that can form the logical foundational basis for the theory of human reasoning and intelligence. Aristotle's quest for a universal rationalization of reasoning can be based on his notion of dialectic argument.

- Dialectic Rationality based on Dialectic Argument offers a new form of science, which we could call Dialectic Science, that goes beyond the traditional Classical Science based on demonstrative necessity.

- Dialectic Science is driven by the need to recognize the complex multi-dimensional nature of real-life problems and looks for holistic and habitual solutions, in the manner that Aristotle understands the problem of ethics in human life.

- A computational view of Aristotle's study of practical ethics can serve as a paradigmatic example for the design and development of AI.

- Dialectic Science builds on dialectic argumentation and dialectic rationality to address ethics in the design of AI via a form of normative guidance, captured through a habitual learning process of continued adaptation for satisficing holistic solutions in the many dimensions of the problem.

- More generally, Dialectic Science aims for sustainability in the role of science to drive the development of society, away from polluting extremes, under a new enlightenment of humility on the limits of the power of human reason.

Aristotle amongst his many concerns he is a practical philosopher with an original concern on how to reason usefully in our life. As such, this could offer wise reflection for the future of science: a future that is less polluting both in our natural environment and in our individual mental and social cosmos.

A new reading of Aristotle from a modern perspective could be instructive for the development of AI. It is useful to return to the original systematic study of human reason and see the dream of reason afresh, for perhaps a new path to understanding the human capacity to think and how this can affect and shape the future nature of humanity.

**Introduction**

This article is about Artificial Intelligence (AI) and Aristotle. It aims to explore some of the past and potential future influence of his work on the developments of AI. This exploration of Aristotle under the lens of AI will be mainly confined to his work on reasoning and human thought and how this could relate to the strategic design and development of AI. Other parts of his work, e.g., his works on Metascience or on Ethics, will play a role in our study, but only insofar that Aristotle connects his investigation of these with the capacity of human thought and reasoning.

It is useful to set out from the start what we mean by AI. What is Artificial Intelligence (AI)? A well-accepted answer is that it is the study of intelligence and its realization in artifact non-human forms. Simple as this statement may be it contains an important element, namely that it refers to human intelligence, implicitly requiring that AI is not an alien form of intelligence, but it is in some way human compatible in the sense that it can be connected and integrated within the sphere of human intelligence. We therefore also need to say what we mean by (human) intelligence, and we will take this to be the cognitive function of structured thought or reasoning at large.

In turn this connection with human intelligence raises the important question of "Why Artificial Intelligence, what for? Again, a simple answer is that AI is about building thinking machines to help energize and expand human thought into new horizons. Although this statement is also generally undisputed it is blurred by market-driven rhetoric on super intelligence for ready-made intelligent and perfect solutions by AI systems. We will adopt the position that the dream nature of AI intelligence is a form of co-intelligence [1] rather than super-intelligence [2].

Let us begin by asking, where does Artificial Intelligence (AI) begin? Is it with Turing who realizes that machines, like the modern computer, could perform intelligent computation and hence perhaps exhibit a form of intelligent thought? Is AI born in 1956 at the Dartmouth workshop organized by Professor John McCarthy, widely considered the father of AI, who coined the name of Artificial Intelligence for the workshop, and soon after presented his seminal paper [3] "Programs with Common Sense"? Another alternative is that AI starts with the early cases of calculating machines or more accurately designs of machines, e.g., the Analytical Engine by Ada Lovelace and Charles Babbage, with the purpose of performing some form of what we might consider an intelligent task, usually that of complex mathematical calculations, and modifying itself during its operation to suit the computation at hand.

All these advances are based on or follow a thread of developments of formal universal systems, usually referred to as logical systems, that are meant to capture the process of human reasoning to conclusions as answers or solutions to (any) given problem. We can start[4] this thread with Leibniz's formal system of numbers for representing complex concepts and the universal reduction of questions to mathematical reasoning, thus needing only mathematical calculation to solve problems, as expressed by his "instruction" of "Calculemus". The thread then leads us to the systems of thought of Boolean Logic and then to Frege who in the late 19th century formulates what is now known as Predicate Calculus or Classical Logic as a formal system that unified the various formal languages for mathematical and scientific reasoning established in the different natural sciences. Classical Logic ever since has formed the bedrock of all Science, including the relatively recent development in the 1950s of Computer Science and Artificial Intelligence, although in the case of AI, as we will see below, the link with Classical Logic is shaken.

But we can take the story of the birth of AI further back. If we accept that the essence of intelligence is structured thought the person who proposes that we can study thought as a system is Aristotle[5]. He aims to understand thinking via a general framework of reasoning and how this could universally capture human structured thought. The influence by Aristotle on the aforementioned developments of Leibnitz and others is clear: Aristotle poses the groundbreaking and non-obvious idea that thinking can be systematized and that this systematization can capture universally human thought. This idea that we could find universal structures in human thought is a prerequisite for opening the possibility for Artificial Intelligence, at least artificial intelligence whose form is compatible with human intelligence[6].

This originality of Aristotle, i.e., of introducing the possibility of the existence of a formal system of reasoning, is widely referred to by the phrase that "Aristotle is the father of Logic". This together with the idea that we can study the systematization of thinking widely across its whole spectrum, covering mathematics, scientific inquiry about natural phenomena, but also humanities, whether this is ethics, politics, philosophy, etc., could justify attributing to Aristotle also the fatherhood of the study of intelligence and thus of Artificial Intelligence.

---

[4] Earlier systems such as that of the philosopher and theologian Ramon Llull [4], where concepts are formalized through an elementary language of letters and whose combinations corresponded to arriving/reasoning at conclusions, would have influenced thinkers like Leibniz at later times.

[5] It can be argued that before Aristotle we have the "invention" of writing as an earlier development of AI. Writing is indeed an artificial construct, a reason for which Socrates objected to its use, that acts as a medium for human reasoning and intelligence. It is not though a system of thought that can be operated by an artifact machine, although with the development of LLMs this is changing. Writing is merely a canvas on which to perform thought by some external entity.

[6] It could be said that the recent developments of AI based on Artificial Neural Networks, such as the development of Large Language Models (LLMs) have intelligence without a formal reasoning system underlying it. But their intelligence is just an extensional one. If we accept that we want from AI to be human compatible it needs to be able to map its intelligence into reasoning at the human cognitive level. i.e., although it is not produced by a formal reasoning system it will need to have a representation by such a system – analogously to what happens in the human brain and mind that represents the operation of the brain at a high cognitive level.

Indeed, Aristotle seems to be motivated by a multitude of reasons for coming up with his original ideas on reasoning and human thought: the beauty of mathematics, the harmony of natural phenomena, the quest for knowledge in humans, the need for a virtuous life of an individual as well as a virtuous collective society. He dares to think that they are all amenable to some formal form of reasoning and to some extent there is an underlying general and abstract system of reasoning that is relevant to all of these matters of investigation.

A most interesting fact about the originality of these ideas of Aristotle, particularly that of the study of systematizing human reasoning, is that he attests to this originality himself. The two passages[7] below say (in free translation) that he could not find prior work on this task of a general theory of (valid) reasoning, despite his extensive efforts in looking for such work:

*Ταύτης δὲ τῆς πραγματείας οὐ τὸ μὲν ἦν τὸ δ' οὐκ ἦν*
*προεξειργασμένον, ἀλλ' οὐδὲν παντελῶς ὑπῆρχεν.*

*Περὶ δὲ τοῦ συλλογίζεσθαι παντελῶς οὐδὲν εἴχομεν πρότερον λέγειν*
*ἢ τριβῇ ζητοῦντες πολὺν χρόνον ἐπονοῦμεν.*

These statements have caused some concern, e.g., that they seem to exclude the a-priori idea and work in mathematics of reasoning to valid conclusions by several methods, including that of reductio ad absurdum. Nevertheless, when one compares these methods with the concrete, but quite general, Aristotelian system of reasoning one sees that there is indeed an essential difference in the universality of reasoning of his proposal with the various loosely connected elements of mathematical reasoning. In fact, we can see that Aristotle's general and abstract work on logical reasoning comes to give an umbrella for mathematical reasoning in ancient Greece, notably the later axiomatic formulation of Euclidean Geometry.

---

[7] Normally, quotes from Aristotle are accompanied by an exact reference of where these can be found in the large corpus of his work. For example, the first passage here would be referenced by *"SE183b34-36"*. We will not do this in this article as the purpose for including these quotes in ancient Greek is more stylistic than for their detailed linguistic study.

**Section 1: Aristotle the father of Reasoning**

As mentioned above Aristotle studies for the first time the problem of systematizing thought in a possibly universal framework. He sets a very general and all-encompassing definition of what is reasoning in the following statement:

*"Ἔστι δὴ συλλογισμὸς λόγος ἐν ᾧ τεθέντων τινῶν ἕτερόν τι τῶν κειμένων ἐξ ἀνάγκης συμβαίνει διὰ τῶν κειμένων."*

This is found several times in his writing as it plays a central role in setting the main requirements of what is reasoning and indeed valid or logical reasoning. In a free translation it states that reasoning is an argument (**λόγος**), a reason, for some conclusion to hold based on some information that we take as given, usually referred to as premises. Notably, the statement requires that the argument supports the conclusion so strongly that it will necessarily follow (*ἐξ ἀνάγκης συμβαίνει*) or hold from the given premises: once we accept the premises the reasoning (argument) renders the conclusion absolutely necessary. This is referred to as the **demonstrative necessity** of the conclusion from the premises.

There are two matters of how to build on this foundational concept of reasoning. One matter concerns the question of whether we can identify standard patterns of reasoning, called demonstrations by Aristotle and in modern language deductions, i.e., general patterns of connection between given premises and a conclusion. The other matter concerns the question of what type of information we can take as premises.

Just to think of the first question is a matter of extreme innovation on the part of Aristotle. To think that, amongst the plurality of human reasoning, whether this is in mathematics or the natural or human sciences, about different topics with a different vocabulary in Natural Language, there could be an underlying basis of reasoning that is independent of what the reasoning is about – independent of the content of the reasoning – is indeed a eureka thought. And indeed, as mentioned above, it is a first prerequisite for AI to be possible, because this can then form a system of intelligence to be automated on a machine[8]!

Before going ahead to analyze further this definition of reasoning it is important to stress the central role that its plays in Aristotle's investigation of what is knowledge and how we acquire knowledge about a subject, with the paradigmatic example of a subject matter being the study of natural phenomena, i.e., of science, and hence the question of what is in general scientific knowledge. Aristotle proposes the idea of what has come to be named **Demonstrative Science**, where knowledge about a subject is organized in terms of demonstrations, according to this basic definition of reasoning, where we have secondary knowledge that follows necessarily from primary knowledge of principles about the subject of study and then by repeated applications of this definition we get further secondary knowledge, from earlier secondary knowledge, etc. The important element of Demonstrative Science is that this generation of new knowledge is done via steps of demonstrative necessity, i.e., where we are absolutely sure that the secondary knowledge will hold true from the truth of the primary knowledge, whose statements are themselves taken to hold absolutely or of "unconditional necessity". From the reverse viewpoint, all knowledge about a subject can eventually be strictly reduced to a set of first principles that are taken as absolute and define the subject of study.

---
[8] Modern AI practitioners may argue that this is not needed as a simulation of the brains neural networks is sufficient. But at least at the current stage and foreseeable future all that the artificial neural network AI systems are achieving is a phenomenology about intelligence without the underlying mechanism for high-level cognitive thought and intelligence: i.e., a matter of appearance of intelligence.

This idea of Demonstrative Science is clearly motivated by the mathematical topics of the time, such as Arithmetic and Geometry, with Aristotle setting up the challenge to turn all scientific investigation, at least of natural phenomena into the same type of framework as these mathematical topics. Indeed, the idea of Demonstrative Science is of extreme historical importance to the development of science: it forms the central dogma in science that all scientific theories trying to explain the respective phenomena in their field should do this in this manner of demonstrative science of Aristotle. In modern theoretical science, this dogma manifests itself by the simple fact that all theories are expressed in the strict language of mathematics. To put it simply, the idea of Demonstrative Science of Aristotle is nothing else than the mathematization of science and the maxim of reductionism in modern science.

**Subsection 1.1: Aristotle's Reasoning System**

Aristotle follows this original idea of the possible existence of structure in thought or reasoning and proceeds to propose a system of patterns of reasoning, according to his basic notion of syllogism presented above, in what has come to be named as Aristotelian Logic or Syllogistic Logic. He develops this system of reasoning based on general and abstract statements, as for example the statement "All As are Bs" where A and B are parameters that can be substituted for specific terms depending on the content of reasoning that we wish to capture, e.g., we could have "All Cats are Four-legged ", "All Squares are Four-sided" or indeed "All Bananas are Four-legged". This abstract parametric nature of the statements gives them their generality, i.e., the potential to capture a wide range of knowledge and thus reason widely in a reasoning system that is based on them. The other important observation of Aristotle that is implicit in the formal language of his syllogistic logic is the fact that despite the plethora of words in Natural Language that we use when we reason what matters most for logical reasoning is only some small set of keywords, like the word "All" in the above statement.

Aristotle then gives his syllogistic system of deductive reasoning as a system where from two statements as the premises we can deduce that a new statement will necessarily follow. For example, if we take as given the two sentences that "All Cats are Four-legged " and "All Four-legged are Runners" then we can reason to the conclusion that "All Cats are Runners". Note that this pattern of reasoning is the same one used to deduce that "All Squares are Four-angled" from the premises "All Squares are Four-sided" and "All Four-sided are Four-angled". And similarly this pattern will allow us to conclude that "All Bananas are Runners" from the premises of "All Bananas are Four-legged" and "All Four-legged are Runners". This latter case of deduction would be a valid argument – i.e., a valid form of reasoning – just as it is for the other two cases. Indeed, human thought has evolved so that if we accept any two pieces of information in the form "All As are Bs" and "All Bs are Cs" the conclusion "All As are Cs" is inescapable, irrespective of the content of A, B and C. What matters is only the form of the statements and the pattern of the connection between the given statements as premises and the statement of the conclusion for this to be a valid case of reasoning.

**Subsection 1.2: Aristotle's Reasoning Universalis**

These original ideas by Aristotle, namely that we can formulate an abstract parametric space of knowledge on which we can base a system of reasoning, and that this space of knowledge can be delimited by a small finite number of structures that give basic patterns of reasoning, form a prerequisite for AI: they open the possibility to form a system of structured thought that can be automated on a machine.

But this alone is not sufficient to form a basis for understanding intelligence in its entirety and generality, unless we dare to think, as Aristotle has done, a further original idea that we can possibly define such an abstract system of reasoning that would be complete, in the sense that all possible deductions (i.e., logical connections that comply with the basic definition in the statement above of what is reasoning) would be expressible and derivable in such an abstract logical system. In other words, Aristotle sets the quest for a system of reasoning such that any case of structured thought by humans as expressed in Natural Language, i.e., not simply mathematical reasoning, would be captured in such a single abstract logical framework. This is the **quest** for **Reasoning Universalis**.

Even though his specific proposal of Syllogistic Logic as such a universal reasoning framework does not meet the desire for completeness, the idea that we can set such a quest for Reasoning Universalis is an extremely important scientific idea, second to none, except possibly the idea of evolution. This reductionist idea of the existence of and quest for a system of reasoning to which all reasoning can be reduced, is the idea that starts the thread of developments that leads us through the two millennia to modern Science and today to AI, as we have described above when answering the question of the birth of AI.

We can justifiably assume that it is this quest for **Reasoning Universalis** together with the idea of Demonstrative Science that comes to drive Leibniz to formulate his system of Mathesis Universalis, with the aim that all reasoning can be reduced to mathematical calculations, and later Frege to formulate the framework of Predicate Logic as a system of reasoning that indeed has a universal nature to underly all scientific knowledge. Frege's predicate logic, which was often characterized as "doing Aristotle's Syllogistic Logic properly", was considered that essentially settles the quest for reasoning universalis in the scientific domain, taking the name of **Classical Logic** to indicate this success. In other words, Classical Logic can be seen as bringing to fruition Aristotle's quest for Reasoning Universalis and Demonstrative Science.

For example, the theory underlying scientific reasoning of Newtonian mechanics, often referred to as **the** Calculus, is nothing more than a specific reasoning system under the universal framework of Classical Logic, where the axioms or premises of the system are the first principles of calculus. In other words, although we usually denote the mathematical calculus as the theory behind modern physics, behind calculus, and indeed behind any other scientific theory, its operation through reasoning patterns of Classical Logic, the logical system that has come to complete the original work of Aristotle with his system of Syllogistic Logic.

After this success of Frege with Classical Logic the story for the quest of reasoning universalis continues, within the restricted domain of mathematical and scientific reasoning. Hilbert a famous mathematician of the early 20$^{th}$ century sets the open challenge to prove that indeed all mathematical problems can be decided upon by a reasoning path, a mathematical proof, as conclusions from the first principles of the mathematical area in which the problem is posed[9].

But this was shown to be impossible by Gödel in his seminal incompleteness theorems. These theorems say that Hilbert's wish is not possible, i.e., there is no general way to reason to the truth or falsity of all statements that we can pose within some area of discourse such as that of the area of mathematical arithmetic[10]. This result of Gödel is justifiably regarded as the most significant mathematical result of the 20$^{th}$ century. Probably, the most important reason for this accolade, is the fact that it was in the effort to understand and replicate this result that Turing formulated his theory of computation through an abstract machine, the Turing Machine, which in turn gave birth to Computer Science.

---

[9] Note that this is a slightly different quest for reasoning universalis: instead of looking for a framework where all reasoning can be encapsulated, we are looking for a framework within which there is always a reasoning path that will take us to a statement that holds true. This pre-supposes that we have an independent way of knowing if something holds or not without knowing why and we are looking to confirm this with a reasoning path or a proof. The analysis of this is beyond the scope of this paper.

[10] Goedel's incompleteness theorems can be illustrated by the following simple example, whose idea comes from Epimenides in Ancient Greece/Crete, called the liar's paradox. If I say that "I am a liar" or that "I am lying" then it is not possible to decide whether I am a liar or not, or that indeed I am lying or not. Another example, is the following question-answer: "Are you happy? I am not unhappy." Again, here it is impossible to decide if this person is happy or not: if we assume that s/he is happy he would have answered "yes". If instead we assume that he is not happy, i.e., unhappy, then her/his answer contradicts this.

What is important to realize is that Turing's Machine is indeed a truly universal framework under which we can reach conclusions in any problem domain with any language of discourse. In this sense it can be seen to satisfy the ultimate quest for Reasoning Universalis of Aristotle as within the universal Turing machine we can reason to (or, as we otherwise say, compute) any conclusion (or solution) that can be possibly reasoned to. Put simply, it can be seen as a reasoning framework that can capture any possible form of reasoning in any field of study. In some sense, the statement that "Classical Logic is the Calculus of Computer Science" indicates this universality par excellence of Computer Science, i.e., that indeed the underlying reasoning framework for Computer Science could not be confined to some specific mathematical framework but it needs to be the whole system of classical reasoning underlying modern science.

There is though a strong caveat to this conclusion of the universality of Computer Science. The Turing machine indeed has a universal nature[11]. But this universality is achieved at a very low level of uniformity – at a mechanistic level – that is far remote from the high cognitive level at which we, as humans, reason, whether this is mathematical reasoning in Classical Logic or everyday common-sense reasoning in Natural Language. In other words, it is not reasoning at the level as Aristotle meant it or would have wanted it, i.e., based on a set of accepted first principles and where the reasoning could be explicated to a human audience! Reasoning that would form a corpus of knowledge that would be understood by humans so that they can built upon it new knowledge etc. For this it is important to know not only the result of reasoning, like the output of a computer which gives us simply the result or a solution to a question, but also to know the details, at some degree depending on our interest, of the reasoning that gives the result or solution. In Computer Science this explication of the reasoning or the process with which we get to the result is not important for two reasons, (1) because when we are studying or using Computer Science, we are not typically interested in building a theory of things and (2) the results of computing are demonstratively necessary, i.e. there are absolute guarantees that they must follow from whatever information we take as given, namely the computer program and the input that leads to the result. This absolute certainty of their results does not encourage the need for an explanation of the reasoning that leads to them.

---

[11] Note that computers are used universally – they do not have a specific purpose when they are built, but can be used for any purpose, i.e., for any reasoning problem that we might be interested in.

## Section 2: The "Science" of Artificial Intelligence

Computer Science opens the way for Artificial Intelligence: it gives an artifact, the computer, on which we could possibly realize a theory of intelligence and thus give us AI as it was original conceived by Turing, McCarthy, and others in the 1950s. The computer has a viable non-human artifact form on which to realize AI and hence the challenge remains with the other half of the definition of AI, namely, to understand or formulate a theory of intelligence in a way that this is computationally realizable. We could then build computers as thinking machines.

How would we approach this challenge? What are the basic principles on which we could develop an understanding or a science of intelligence? Is intelligence, and hence also AI, a subject of a Classical Science that follows the long tradition of scientific theory development over the centuries as we saw above? With regards to Aristotle is AI a Demonstrative Science? What would Aristotle think of AI[12]? What would Aristotle, as the original scholar of intelligence with his pioneering study of structuring and systematizing thought, consider as the science of intelligence? Would Aristotle consider that Intelligence, and hence also AI, was a subject of demonstrative science? These questions are closely connected to whether Aristotle thought that fields of investigation that today we call the humanities, should also be studied as fields of demonstrative science?

To get a glimpse into this we can go back to his central definition of a syllogism and note that this will have guided Aristotle both when he is studying reasoning itself but also when he is applying reasoning to science, social problems, and philosophy. Within this plurality of concerns, he seems to realize that there is a need for two cases of reasoning or of logical demonstration according to, one could say, the strength of the premises. One case is where the premises are **absolutely undisputed**, as for example in 2D geometry with premises like «there is only one straight line connecting two different points on a plane», and hence the conclusions drawn from such premises are also so – this is referred to as unconditional demonstration. The other case of reasoning is where the premises cannot be taken as statements of absolute necessity, but they have a merit of universality which is described by Aristotle as "hos epi to polu" (ὡς ἐπὶ τὸ πολύ) statements. One way that this has been translated by scholars is **"for the most part"**, e.g., the statement of «human beings act from their own conception of the good» is such a "for the most part" premise in the study of ethics.

Under such premises demonstration is softer than in the first case. We can call this plain demonstration as opposed to unconditional demonstration and indeed with plain demonstration this may need to be qualified in order for the conclusion to hold. The important observation by Aristotle is that if we want to apply reasoning outside mathematics, notably for science or ethics, then we need to consider premises and demonstrative reasoning of this second "hos epi to polu" kind.

---

[12] Aristotle, like others is ancient Greece had considered that some human activities could be carried out by artificial beings – they could be artificialized. But did they also think of artificializing intelligence or reasoning? Aristotle, like Plato, concerned himself with the question of whether intelligence (noesis) could exit independently of the hardware of the human brain and/or human body.

Enter modern AI. In modern AI terms we would recognize these two cases as those of non-defeasible and defeasible knowledge and reasoning respectively. For the case of defeasible knowledge statements are qualified either explicitly, or implicitly within the defeasible reasoning that we apply with them. For example, suppose we are given the statement "Every morning I have a shower at home.". For the question "Did I have a shower this morning?" we would consider the statement as one expressing an absolute necessity and answer "yes". But in a separate case of reasoning where we are also given the facts "The water supply was cutoff last night and has not come back.", the statement loses its absolute necessity and the demonstration of the answer yes to the same question fails. This phenomenon has been observed empirically in common sense human reasoning in Cognitive Science experiments [5] which generally go under the name "the suppression task", to reflect that the answer is suppressed under additional information.

In general, what is happening here is that some additional relevant information would **qualify** the statement from a necessary one to a default one. Defeasible statements are thus default statements which necessarily hold for the most part but in special contexts they break down. We are then faced with the qualification problem [6] to know and express these contexts for each defeasible statement either as explicit qualifications in the statement, e.g., "Every morning I take a shower unless there is no water supply" or implicitly in the defeasible reasoning. This was recognized as an extremely important problem at the start of modern AI in the 60s and at the same time extremely difficult to scale up to a large corpus of knowledge. We will return to it in the next sections.

**Subsection 2.1: The early history of modern AI**

Before doing so, let us continue the historical thread of development of science with the birth of modern AI in the 1950s. The modern pioneers of AI were concerned with artificializing our general intelligence, not simply the specific and narrow formal mathematical intelligence. Indeed, at the very beginning of the endeavor for AI there was a strong link with Cognitive Science, which studies human intelligence in the large [7]. The study of intelligence within Cognitive Science is carried out more at the phenomenological level with descriptive models of thought and intelligence rather than the development of predictive theories of thought. It was then hoped that using this phenomenology we could find underlying scientific theories that would account for the phenomena of cognitive thought and that they were computational realizable. Such a computational theory could be a theory of logical reasoning that would underlie cognition in the large.

But it was immediately clear that such a logical theory for cognition could not be that of Frege's Classical Logic. In other words, AI could not be a Classical Science! The modern dogma that any scientific theory is a theory within the framework of Classical Logic had to be broken. Nevertheless, the early AI community (implicitly) decided that a theory of intelligence would still need to be a demonstrative science in the sense of Aristotle. This decision was driven by the "culture of science" of following the trusted logicism and reductionism of science. We would just need to find another logic different from Classical Logic but one which again would be used in the demonstrative necessity format as set out by Aristotle. The culture of (theoretical) science was so strong that it resisted being abandon even when its protagonist of Classical Logic had to be abandoned!

The fact that there is a disparity between formal logical reasoning and common-sense everyday reasoning, usually referred to within Cognitive Science/Psychology as human reasoning, as observed in the general human population was already well known by the start of the modern study of AI. Several experiments [8] confirmed that humans are not formal logical reasoners: the correlation between human reasoning and formal logical reasoning under Classical Logic is not very high[13]. But it is also not very low. These empirical findings together with the observation of the simple technical characteristic of human reasoning that conclusions reached can be withdrawn in the face of new information, a situation that cannot happen with Classical Logic, drove a search for new logics, generally called non-monotonic logics (NMLs). This name was given exactly to reflect this basic technical difference that their conclusions are not monotonic, like those of classical logic's, whose conclusions once they are established, they remain so irrespective of what additional information is added to the theory. For example, mathematical theorems once proved within a framework they continue to hold irrespective of any additional premises that we might add in the mathematical framework. Theorems are forever, and hence we see for example Pythagoras theorem surviving the passage of millennia[14].

As already mentioned above, from the very the start of modern AI we have an abandonment of Classical Logic as the framework on which to build a theory of intelligence, and a search starts for non-monotonic logics to model human reasoning as demonstrative reasoning under these new logics. We have for example the logical framework of Circumscription, proposed by McCarthy, as a logical basis for commonsense reasoning with particular interest on how humans reason about actions and the change they bring about in the world. In this we are trying to formulate a theory for the human intelligence of reasoning about information over time, such as how we reason to new information by persistence over time, e.g., from the premise/fact that "Bob is at work early in the morning" to the conclusion that "Bob is still at work at midday", or from the additional fact that "Bob had left for the hospital at 11am" that "Bob is not at work at midday", etc.

---

[13] This has continued over the decades of the 20th century with many experimental studies within Cognitive Psychology but also outside this field, where for example Richard Thaler is awarded the Nobel prize in Economics in 2017 for his work behavior economic where humans do not behave as rational reasoners in their economic decision making.

[14] Note that theorems can stop holding if we take away some of the premises under which they hold but as said above, they never fail if we keep adding additional premises.

Similarly, we have had other proposals of non-monotonic logics which aim to formulate a theory about our knowledge and beliefs so that they can capture our intelligent reasoning about what we know. For example, so that we can reason from the fact that "I know that Bob knows that Mary is at the office" to the conclusion that "I know that Bob can warn Mary of the danger that I believe she is in" and from the additional fact that "I know that Bob knows that I know he is with Mary' to the conclusion that "Bob would be upset if I did not contact him to tell him about the danger".

A large plethora of different non-monotonic logics appeared all searching for the holy grail of a formal theory of human reasoning. The belief was that "all we need is to formulate/find the right logic. Then following its reasoning would give us human reasoning".

But the tension between the defeasibility nature of human reasoning and the absolute nature of (Aristotelian) demonstrative necessity that was imposed on the search for the "right logic" just meant that more and more different proposals for such logics were proposed. These logics needed to have a flexible inference, in the same manner as human do when drawing inferences, to missing or ambiguous information and tolerant to (apparently) contradictory information. They were/are aimed to capture defeasible reasoning, i.e., reasoning that could turn out to be wrong, and yet their design to do this was via some form of absolute necessity! This made it very difficult, over four decades, to deliver on their promise of "AI systems with common sense" and "human-like natural intelligence of AI systems". To ensure the absolute certainty of conclusions is a complex and difficult task especially in a dynamic setting where conclusions could be withdrawn. For example, we would need to have the right belief revision mechanism [9] that would take us from past held conclusions to news ones. This was studied extensively in parallel to the development of NMLs, but again under a framework of demonstrative necessity that would give the new absolutely necessary revision results. But why strive for necessity when it is not generally there? Something else was needed to capture the flexibility in reasoning.

Then in the early 1990s, Bob Kowalski at Imperial College, London, made the connection of non-monotonic reasoning with argumentation, first within the non-monotonic framework of Logic Programming, a flagship of AI for several decades, which he had founded together with Alain Colmerauer in France. It soon became clear to the community that essentially all proposals of NMLs could be reformulated within the framework of argumentation-based reasoning [10,11,12]. This AI approach to argumentation, sometimes referred to as Computational Argumentation, was also motivated and to some extent grounded on earlier foundational work [13,14,15] on human argumentation in Philosophy and Cognitive Science.

This result of reconciling non-monotonic logics through argumentation has meant that we could now have formalizations of aspects of human reasoning, such as that of reasoning about actions and change [16,17] which, as mentioned above, was at the heart of the modern start of AI, and that would allow us to build computational models of narrative comprehension akin to the way humans perform this task [18].

At the same time, within the separate community of Cognitive Science, studies were showing the close link between argumentation and human reasoning, see for example the book "The Enigma of Reason" [19]. It then became apparent that we could synthesize a framework of computational argumentation informed by cognitive principles of human reasoning to obtain a framework, called Cognitive (Machine) Argumentation, as a suitable framework to model human reasoning in its various forms. Within this framework it is possible to offer an alternative uniform view and reconcile the empirical results of the many classical studies of the logic of human reasoning within the area of the Psychology of Reasoning of Cognitive Science. Indeed, it has been shown to capture well the human empirical data from several different experiments that are traditionally used in Cognitive Science to evaluate cognitive models of human reasoning [20]. These empirical evaluation domains include the Wason ``Selection Task'' [21] where humans are tested on the way they reason about conditionals and the ``Suppression Task'' [8] where the non-monotonic nature of human reasoning is observed. Interestingly, they also include human ``Syllogistic Reasoning'' with experiments on how humans reason on the original Aristotelian syllogisms (with argumentation winning an international challenge in 2017 for the closest model of this human form of reasoning).  [22]). Cognitive argumentation accounts uniformly for the empirical data in these domains in a cognitively adequate way that also reflects well the variation of reasoning across the human population.

And to complete the "dominance" of argumentation as a reasoning framework it is also possible to show in very precise terms that (strict) demonstration as it is formulated within Classical Logic, can be equivalently captured as a special case of argumentation-based inference. Indeed, we can define a formal logic, called Argumentation Logic (AL), as a realization of the same general computational argumentation framework that underlies all non-monotonic logics and the empirical data of human reasoning behaviour, and show that this captures precisely classical deductive reasoning [23]. This means that argumentation is now a strong candidate for Reasoning Universalis across the whole spectrum of human thought, irrespective of the topic and content of reasoning, whether in mathematics, science, humanities or in common sense everyday reasoning.

So, what is argumentation-based inference? For that we can go back to Aristotle, the father of systematic Dialectic Argument! But before doing so, let us have a brief look into argumentation-based inference, as it is found in today's AI, through an instructive and important historical example.

**Subsection 2.2: An AI example of Argumentation-based human reasoning**

Let us present a little of the technical detail of how argumentation can form the logical foundation for human reasoning - argumentation as reasoning universalis - and as such a theoretical framework for formulating human intelligence. We will do so via the specific and important example of how argumentation captures human reasoning in time about the effects of actions and the change they bring about in what we know of our environment.

The argumentation-based theory for this problem is quite simple in comparison with many of the non-monotonic logics that have also been proposed for the same problem. We use three types of arguments. Persistence arguments, where from the knowledge that a property holds at some time, we can support and conclude that this property also holds at later times, e.g., if we know that the light is off in the room in at dawn then we can use these arguments to support the claim that the light is still off at mid-morning or at midday or any time after. Another time of arguments are called generation arguments, where from the knowledge that an action occurs at some time we can support that a property that it is generally initiated by the type of action will start to hold at that time (or if we want to be pedantic, start at a time immediately afterwards), e.g. if we know that someone turn-on the light switch at some time in the mid-morning then we can use the generation argument to support that the light is on at midmorning (and then by the above persistence argument that the light is on at times after midmorning). Finally, we also have qualification arguments, where from the knowledge of some information we can support that despite the occurrence of an action its effect will not hold – will not be generated, e.g., if we know that there is no electricity in the building at the time of turning the switch on, then the action of turn-on the light switch will fail to bring about the effect of the light on. (We can also have qualification arguments that support that an action cannot possibly happen).

Reasoning in this theory – as in any argumentation theory – occurs via debates between these three types of arguments, a debate that we can imagine occurs in the human mind when we take a decision of what holds in the world at a certain time. In an argumentation debate what plays a crucial role is the relative strength between (opposing) arguments. In our case here, generally, generation arguments are stronger than persistence arguments and qualification arguments are stronger than generation arguments. Furthermore, any argument that is based on information that is later than the information of which another (opposing) argument is based, is stronger. Hence, in the example above where we know that there was an action of turn-on the switch at midmorning, we conclude that just after midmorning the light is on because the generation argument is stronger than the persistence argument from the early morning, but if we know that there was no-electricity at the building in the morning we will not conclude so but conclude that the light is off based on the persistence argument for non-electricity from morning to midmorning and the qualification argument on the action of turning-on the switch.

Let us suppose we are then told that the electricity came back on sometime in the morning but not if this was before or after midmorning. Given this extra information we will not be able to conclude if the light is on or off after midmorning because we will have an equally strong case of arguments to support either conclusion. For the position of light-on its argument case contains the belief or assumption that the electricity came on before midmorning, together with the persistence argument of electricity till midmorning and the generation argument from the action of turning on the switch. For the position of light-off its case contains the opposite but equal in strength assumption that the electricity came on after midmorning, together with the persistence argument of no-electricity till midmorning and qualification argument on the action of turning-on the switch.

In such cases, the argumentation-based reasoning recognizes the lack of information to demonstrative decide and accepts that the reasoner will be in a dilemma with different cases that are able to explain each of the possibilities. Indeed, with argumentation as the reasoning inference engine, explanation comes for free as explanations can be essentially gathered off the debate between arguments and counter-arguments that led to a winning or possible conclusion. This means that these explanations can contain both information for why to choose the conclusion - the explanandum, but also information for why not to choose other conclusions.

The argumentation theory described above is indeed quite powerful in capturing human reasoning with time, actions, and change by action occurrences. It can also form the basis for modelling how humans plan to bring about a desired situation through a sequence of actions that will form the premises of arguments that together make up a strong case for the state of the world we are trying to reach.

Interestingly, when we look into this argumentation theory, we can see an analogy with Newton's theory of motion in Physics. Persistence arguments essentially capture a sort of inertia in our knowledge of the state of the world, analogously to the inertia law of Newton and generation arguments reflect, analogously with the second law of Newton, the fact that the state of affairs will change over inertia by some external influence – although Newton's law tells us more about the change, giving us a demonstrative conclusion of what the change will be amongst the several possibilities (provided we know the influencing external force, i.e., the action, at an appropriate level of detailed.).

**Subsection 2.3: Argumentation in real-life AI problems**

A natural question to ask is whether this argumentation theory can form the basis for developing real-life AI applications where such reasoning with time is essential. The answer to this question is no, or at least not yet[15]. This is because to apply a theory in the context of a real-life problem a lot more is needed than the theory itself. We need tools, typically these are mathematical tools, that would allow us to reason with the theory in the complex situations of real-life problems – sometimes even specialised tools for the specific type of problem that we are interested. We can have the principles, but we also need to know how to do the reasoning based on them to decide about questions/conclusions that we are interested in.

This is not new to AI. It is true for any scientific theory. For example, with Newton's theory of motion, even when we try to apply this to a simple (realistic) 3-body problem of how three bodies will affect the motion of each other, we need to develop a heavy specialized mathematics machinery to help us work out approximations to the solution of the problem. This effort to develop mathematics **to engineer** solutions out of Newton's theory, was started by Newton, but continued for centuries so that we can reap results of reasoning from the theory. Note also that in almost all cases this engineering mathematics are approximations, not exact representations of the theory, and this is also because theories are formulated at idealized levels which are not found in real-life problems. The complexity of real-life problems is so hard that we can only have approximations of what the theory tells us about them.

For the case of argumentation-based theories for human reasoning in AI, as the example of reasoning with actions and change that we saw above, there has not been any developments to provide methods[16] for reasoning to solutions, analogous to the methods of engineering mathematics, that would allow them to be applied in real-life practice. In effect, what we need is the analogue of engineering mathematics that would recognize the contextual relative strength between arguments that are involved in the reasoning – debate of arguments for and against. Methods that would recognize necessary conditions over sufficient conditions, or specific exceptions over general cases, or graded modality of statements from necessity to likelihood and many other such strength comparisons captured in the rich expression of natural language.

Such "engineering mathematics" was missing in AI, except possibly that of Argument Mining [24] methods for the automatic identification and extraction of the argumentative structure of inference within Natural Language. Today the development of Large Language Models (LLMs) can in fact be seen as an analogous approximation method to engineer solutions to reasoning problems. We can see today how we could use LLMs to allow us to apply argumentation-based theories in real-life problems. For example, in reasoning about actions and change, the LLMs can provide approximately the underlying connections between actions and their effects as well as the information of what information constitutes a qualification on the different connections of actions and their effects. Viewing LLMs as engineering methods for a theory of human reasoning and bringing them together with the theory can form a solid platform for the application to real-life problems[17]. The connection between argumentation theory and LLMs could actually be deeper, where the former is a good theory or a view model of the operation of LLMs, despite the fact that LLMs are build and operate internally in a very different way [25]. Interestingly, when comparing the behaviour of an argumentation-based reasoning, called COGNICA, with that of an LLM system, these coincide well in their answers, i.e., on the conclusions drawn, but can have noticeable differences in the explanations that they provide for their results, i.e., they can differ in their description of their reasoning that supports their (same) conclusions.

---

[15] Note that even the Debater system of IBM, that has been around since 2018, is not able to fulfil the needs of real-life reasoning at the human level.

[16] There have been attempts to do so, e.g., by knowledge and concept graphs that aim to catalogue in some way the vast mundane pieces of information about the world we live in that we hold in our minds, but these have only a limited success, perhaps because they are not designed as approximations, but are still tight to a strict absolute representation of non-strict (default – for the most part) common sense knowledge.

---

[17] The latest version of COGNICA, a system implementing Cognitive Argumentation, does this by working in a hybrid way where argumentation is sandwiched between an LLM. The LLM model deals with the natural language interface matters and the low-level common sense background knowledge that might be needed in the argumentation-based cognitive reasoning.

## Section 3: Aristotle, a protozoon of AI?

Why did Aristotle decide to set reasoning as a field of systematic study? What was his motivation for this when no one else was doing so? And why within this new field of his, he sets the quest for Reasoning Universalis that would cover all human reasoning across the board?

The answer to these questions, although not explicitly given, is prevalent in his work: because he wants to apply or connect or relate rational reasoning generally in human affairs. Aristotle did not keep his work on logical reasoning simply for mathematics. Quite the contrary, he relates this to his study of natural phenomena, human behaviour in politics, rhetoric, ethics and his work on philosophy. He wants to have a general understanding of reasoning and human intellect that he can use in his studies across all fields. **This makes him an original intellectual of AI, if not the protozoon of AI.**

Would Aristotle have stopped at the study of reasoning simply for the sake of reasoning and did not involve himself with the bigger question of how logical or rational reasoning relates to other issues in life, then there would be no strong reason to connect him to AI. Instead, he conceives the "rational self", a part of any human that has a very high value and significance in life. Reasoning can regulate many parts of our individual and social lives and Aristotle encourages us to become as rational as possible – as good reasoners as possible.

This interest of applying or relating reasoning to questions in other disciplines can be seen hiding in his central definition of reasoning given above and copied here:

*"Ἔστι δὴ συλλογισμὸς λόγος ἐν ᾧ τεθέντων τινῶν ἕτερόν τι τῶν κειμένων ἐξ ἀνάγκης συμβαίνει διὰ τῶν κειμένων."*

In this, it is stated that the conclusion of the reasoning, "*συμβαίνει*", which means that it follows from the premises, but not only as a piece of information. It also says that it will be part of the world about which we are reasoning, i.e., that we will have a way, external to the reasoning, to confirm this. For example, from the premise that a certain two-dimensional shape is a triangle, we can reason to the conclusion that it has three angles, and we can independently confirm this (if we wished) by observing the triangle. Similarly, the conclusion that "Socrates is mortal" that follows from the premise that "Socrates is human" is something that can be confirmed in our biological world. Some scholars have also suggested that Aristotle meant this connection to be causal, i.e., in the world we are reasoning the premises cause or bring about the conclusion, leading to his idea of Demonstrative Science, at least for natural phenomena. In our case, it is sufficient to assume that the choice of this expression, of the word "*συμβαίνει*", is another indication that he is not developing a theory of reasoning in vacuum, but so that it can help him develop his theories across a whole spectrum of topics.

At the center of this desire to apply reasoning across the board is his quest for Reasoning Universalis and whether this can be satisfied by his syllogistic framework of Demonstrative necessity. Can any field of study be cast into Demonstrative Science? Aristotle new that there is an inherent challenge here, namely that reasoning applied outside the close realm of mathematics could not always satisfy the strict requirement of demonstrative necessity. He understands that alongside the strong demonstration form of reasoning where conclusions follow with certainty, there is the need for a reasoning form where this is not so but that the conclusion holds in a weaker way: it is strongly plausible but not absolutely or it holds for the most part but not in any possible case.

This is the distinction of non-defeasible and defeasible reasoning that we saw above, where modern AI aimed to formulate logical frameworks for defeasible reasoning. Let us see how Aristotle tries to formulate this weaker form of defeasible reasoning. He tries two things. The first one is to introduce the idea of modality statements so that statements can be differentiated as strongly necessary ones, i.e., "necessarily P" or weaker ones "possibly P" for some statement P, where the latter ones are only plausible but not necessary. Aristotle attempts to define a Modal Syllogistic logic [26] in analogy to his Syllogistic Logic (or Assertoric Syllogistic Logic as it is called when we want to explicate the difference) where syllogistic statements can contain modalities, e.g., we can have "Necessarily all As are Bs" or "Possibly no As are Bs", etc. He tries, like in the assertoric case, to determine which inferences are valid from pairs statements taken from assertoric and modal statements. For example, he would claim that "Necessarily all Bananas are Yellow" and "All Yellows are Bright" gives a valid inference of "Necessarily all Bananas are Bright", whereas the premises of "All Bananas are Yellow" and "Necessarily, all Yellows are Bright" do not imply that "Necessarily all Bananas are Bright". This asymmetry was not easy to reconcile, and many scholars find it difficult to follow and indeed accept Aristotle's treatment of modal logic [27].

Despite the quite heavy criticism of Aristotle's treatment of modal logic, when we view this within his wider concern of how reasoning relates to human thinking, outside the strict realm of abstract (e.g., mathematical) reasoning, we can hypothesize that he does this in recognition that something more is needed for reasoning in the large[18]. He recognizes that not all statements are the same – they do not carry the same weight in reasoning.

The modalities that he introduces are a means to capture this variation in the strength of statements. Of course, modalities are present in Natural Language and again he picks them so that he can extend his framework of systematizing reasoning. But his modal-logic based attempt, as well as those of the modern elaborated theories of modal logic, to address the wider question of formalizing human reasoning, are inadequate for the task[19]. Nevertheless, the important observation is that the use of modalities highlights the fact that there is a relative strength between statements when we reason in Natural language and modalities is a way to explicate this strength.

---

[18] When we refer to reasoning in the large, we can think of this as the human reasoning that we carry out in Natural Language.

The existence of statements of different strength is widely recognized by Aristotle[20]. He recognizes that we can have statements that we may want to take as premises in our reasoning but where these statements are not of an absolute nature. Such premises are not of unconditional necessity. They are statements which are potentially (δυνάμει) necessary or in the modern AI terminology, default statements. This can occur both in investigations within the field of science where enmattered statements in some scientific field, e.g., in the taxonomies of biology, cannot be universally true, or in fields of investigation that are not directly connected to nature, such as the field of ethics, where it is more likely to have premise statements that are not unconditionally necessary.

Hence, we have a tension between requiring that we have demonstrative necessity of conclusions and the fact that the premises themselves are not of unconditionally necessary. We therefore need to find a way to reconcile these things. Aristotle appears to try to address this, by identifying another type of "hos epi to polu" statements. In free translation, these are statements "holding for the most part" (from Aristotelian scholarship) or "holding almost always" (borrowed from measure theory in mathematics). Such statements are not rare, but can be found across many of his studies, e.g., in biology, ethics and rhetoric and indeed in his metaphysical study of knowledge. Nevertheless, it seems that a coherent and comprehensive investigation of reasoning with such statements is lacking in Aristotle and somehow this has persisted in the ensuing future investigations of the nature of reasoning.

Until today, where, as presented above, we have the study of defeasible reasoning in AI based on argumentation as the foundation of reasoning. In this approach we accommodate non-exact statements by accepting their inexactness and allow the reasoning to be non-exact, in the sense that it can result in a dilemma: an informed dilemma where the choices are reasonably justified. But this form of defeasible reasoning is non-other than that of dialectic argument of Aristotle as he had studied it within the context of a debate that we find for example in rhetoric or politics. In modern argumentation, the notion of Aristotle's Dialectic Argument can be seen to be captured within different formal frameworks [28, 29, 30] and methods. In dialectic argumentation we need to reason with different opinions as the premises or principles held by each side of the debate, all of which are statements that cannot be unconditionally necessary.

---

[19] The reason for this inadequacy is again the insistence on demonstrative necessity in the extended reasoning of modal logic. For the present article it is not necessary to get into the technical details of this.
[20] The existence relative strength of statements can also be implicit. This is more relevant to Topics and modern AI argumentation.

Aristotle calls these ἑνδοξα, meaning that they carry some form of credibility and hence it is reasonable that they can be taken as the basis, or premises, of arguments to reason with. But clearly, such reasoning needs to be defeasible as it crucially depends on premises that cannot be taken of absolute necessity.

Does Aristotle make this same modern connection between defeasible reasoning and dialectic argument? For example, does he connect his study of reasoning with "hos epi to polu" statements with his work on dialectic argument? It seems not, although we cannot be sure. Maybe this is because this would go against his grand plan of demonstrative science. But why does he not make the connection at least for defeasible reasoning in some cases, like for example in the case of his study of ethics? Or maybe he does so, because what we see at the end as the large picture of his theory of ethics is of a different nature from that of demonstrative science? Reasoning plays a central role in his theory of ethics but yet he considers ethics as a different kind of science, a practical science, as Aristotle calls ethics and other topics of investigation.

This separation as a different genus of science may be because of a different interest from Aristotle at the metaphysical level. But we can also see this difference from the perspective of the need for a different genus of reasoning in this type of problems, the defeasibility of reasoning. A need for different type of reasoning that derives from the real-life complexity of the problem that cannot be simplified in any meaningful way as we could do for other scientific fields. Given the defeasible nature of reasoning involved in such investigations of practical sciences and the link that we saw above of defeasible reasoning to Dialectic Argument we could call this, **Dialectic Science**, as opposed to Demonstrative Science or in modern terms, Classical Science.

**Subsection 3.1: Aristotle's "Computational Ethics"**

Let us then go ahead and see the nature and main characteristics of Aristotle's theory of ethics: what is its essential structure, what is the role of reasoning in ethics and that of defeasibility in ethical reasoning. We will be concerned with these questions in the large, i.e., how they can help shape at the macro level the theory of Aristotle's ethics. Also, in our presentation of Aristotle's ethics we are not interested in what it is said about virtue and morality itself – we are not going to discuss under which morals the theory is based. We are only interested here in its logical and computational nature as a paradigmatic scientific theory for AI and indeed, as we will see below, for any scientific theory across all fields.

Although a computational description of Aristotle's ethics may appear novel it is important to stress that what we will present below is not new in this paper, but rather it follows the state of the art in the scholarly work on Aristotle and Ethics (e.g., as found in the Stanford Encyclopedia Entry on Aristotle's Ethics by Professor Richard Kraut). What we will try to do here is to connect these pieces in a model that could help guide us in the formalization of intelligence for the purposes of AI.

Aristotle's approached and thought about ethics not as a usual scientific theory but more as a study of a life's practice to guide our behaviour. Nevertheless, he has a problem in mind to solve. How to live a virtuous life leading to human well-being. As such we can think of his work as a theory whose proposed solutions would reflect the needs of the problem. We will therefore refer to Aristotle's theory of ethics according to this extended meaning of what is a theory, and we will be interested in this theory primarily as a theory of reasoning.

Aristotle would have recognized that reasoning about ethics and taking decisions that are motivated by moral considerations is quite complex and that it would be difficult to formulate ethics as a Demonstrative Science. They were two interrelated reasons for this: the first is the defeasibility of reasoning and decision making as we have discussed above and the second is that the reasoning necessarily would need to involve several criteria of different form, such as motivational and aspirational but also practical ones.

His realization that the reasoning involved in ethics is defeasible can be seen quite explicitly in his introduction of the notion of leniency in justice[21], where justice here may refer to the law of the land but also to any other context where we are taking some ethically related decision. Aristotle realizes that the law as written down is not an exact statement of absolute necessity and that there can be situations, when we are administering the law, that we can see its failing, where then leniency is needed to be applied. Leniency is applied to correct[22] the failing law and we need this because we know that there will be situations in the future where the law will fail, as we cannot foresee all possibilities. He states this quite explicitly in "ἐπανόρθωμα νομίμου δικαίου", i.e., in free translation, that leniency is a corrective process of the law.

In other words, as we cannot foresee all the qualifications that a law might have – just like the qualifications that we have presented above in common sense reasoning, e.g., that "no electricity" qualifies the "effect law" that the action of "turn-on the switch initiates light" - so that the law can state what to do in such situations, we need to have a process which will apply the qualified reasoning when we realize that we are indeed in such an exceptional situation. This is exactly the same process of human reasoning in the large, where we will apply the law "all birds can fly" as a default law to generally conclude that any given bird can fly but will qualify this in cases where the bird is a penguin, or it is dead, or it is just born, or it is injured, or has lost its feathers, or the bird is space with no air, or … . We use such default laws, without explicitly qualifying them, because they work well for the most part to give us a valid conclusion quickly but when we recognize we are in a special situation we would block the law, unless again there is reason to qualify the qualification, e.g., it is a penguin but it is also a super penguin, or it is dead but it is in heaven, or … .

---

[21] The recent book of N. Paraskeuopoulos [31], "The lenient algorithm: From Aristotelian thought to Artificial Intelligence" (in Greek), is an excellent exposition on this work of Aristotle and its relation to AI.
[22] Aristotle's leniency is not that of sympathy or pity, but it is a purely reasoning mechanism to correct the necessary failing of the written law.

In conclusion, there are two important observations in Aristotle's introduction of the notion of leniency. First that this is actually part of the notion of justice in law and secondly that leniency is a process within the act of implementing the law (or administering justice through the law), a process that we can identify as the same process of defeasible reasoning that we meet in common sense reasoning and other forms of human reasoning at large.

The need for leniency originates from the complexity of the problem at hand where many factors can play a role and thus our knowledge cannot be expressed in a compact and absolute statements. Aristotle realizes the multidimensional complexity of the problem of ethics and that good ethical bevaviour would need to come out of thinking together, or weighing together, many different considerations such as virtue and honor but also personal happiness. At another level, his argues that ethical behaviour concerns a fusion of Logos, Pathos and Ethos, three quite different and often competing considerations. In whatever way one looks at his work on ethics, its characteristic that the problem is necessarily one of multi-criteria decision making is inescapable.

Another important characteristic of his understanding of ethics is that although he indicates that there is an ultimate goal (a telos) to be served by an ethical behaviour, namely that of eudaimonia (pleasure from fulfillment), this does not mean that this criterion should dictate our reasoning and actions. This may sound paradoxical at first, but again the reason for this is the fact that this criterion of eudaimonia has itself a multi-dimensional complex form that is context sensitive depending on the occasion we find ourselves each time we need to take an ethically related decision. Also, despite its central role it does not mean that it is always dominant as we would have situations where other considerations would take precedent, e.g., in the simplest case when our safety or physiological needs become critical.

All this means that a decision theory, or a logical rational theory, for ethics cannot simply be one where we find the action/behaviour that maximizes this central criterion of eudaimonia, or for that matter any other consideration, and follow this. In other words, we cannot simply apply absolute rationality on the central criterion, which in effect would mean that we follow a demonstrative theory, thus turning ethics into an Aristotelian Science. Instead, what Aristotle proposes is that we build eudaimonia in time habitually by learning in time how, where possible, to serve better and better this ultimate goal but not always insisting on a best value for it. This would be a sustainable way to go about the problem where our theory improves over time in serving this criterion cumulatively building better (but not absolute) solutions.

We can imagine that the inescapable multi-criteria nature of the problem of ethics, leads Aristotle to look for solutions that somehow balance the different criteria. Aristotle's quest for the "middle ground", when focusing in any one criterion, is a natural computational consequence of this multi-criteria nature of ethics and indeed of any multi-dimensional problem. Aristotle would have realized that to find a "natural balance", not simply an arithmetic balance, over the many considerations that are involved he would need to avoid the extreme values in each single consideration.

Another way that the complexity of the theory of ethics shows itself is the fact that its principle statements, its premises, and therefore also its conclusions, cannot be taken as statements of absolute necessity. This means that we must be "satisfied by satisficing" solutions, i.e., with solutions that are sufficiently good, as a whole, but also when we look at each relevant consideration separately, we again have a sufficiently acceptable value. This is then simply another expression of his maxim for the middle ground.

The complexity of the reality of the problem points towards the need for something drastically different from the investigation of other fields. That it cannot be addressed (i.e., a theory for ethics cannot be formulated) via a one-shot process at the start of the endeavor of setting down the principles and working from these, as Aristotle would argue for a scientific theory for a part of the natural world. Instead, it is a matter of continuous development of the theory (knowledge) under which we learn to take ethical decisions. He understands that it is futile to expect that one would get the theory right from the start and one would need to keep working on it for hopefully a steady improvement. This is again because of the multi-dimensional character of the problem and the open nature of the reality of the problem where we cannot see a-priori all the different contexts of the problem.

In other words, we need to accept the difficulty of the problem – and that there are no perfect solutions to it – to look for a theory that allows us to adapt and develop in time. As Aristotle would say, we need **habitual solutions**, borrowing the idea/term from Aesop's "One swallow does not bring Spring". From the perspective of modern science this idea would look quite strange as in science we always strive to have a one-shot theory formation (i.e., any proposal that is found wanting form the start would not be permissible) and a change in a theory occurs only when we have a new refutation and most of the times this means a new and probably quite different reformulation of the theory.

This realization of Aristotle to replace a theory of absolute perfection with that of a plane of solutions of sub-optimal performance which improves in time and which sustainably adapts is of crucial importance when we are looking to be informed from Aristotle for the development of today's AI. How do we build AI system? Striving for perfection from the start or for an acceptable and useful performance that can sustainably improve? It reflects an honest admission of the difficulty of the problem and steers us to the second option.

To summarize, in ethics we see Aristotle tackling a real-life problem where reasoning cannot be confined into a well-managed space so that demonstrative necessity can be applied. This is exactly what we have in AI where the problem is also multi-dimensional where for example, we have performance criteria of AI systems competing with other criteria such as, ethical criteria of fairness and transparency. His computational approach to ethics, with its quite different nature from traditional scientific investigation, can therefore inform and guide us for developing a theory of AI. In fact, Aristotle's work on ethics has a double role to play in today's AI: to give us a framework for a theory of intelligence at large – a theory of dialectic science based on dialectic rationality as we will see below – and to help accommodate ethics itself in AI in what has come to be called AI Ethics[23].

Perhaps the most interesting observation from this purely systemic perspective of Aristotle's ethics is the fact that indeed his theory of ethics is amenable to such a computational view. Many earlier philosophers were highly involved in the study of ethics and morality. What makes Aristotle different, is the computational reasoning perspective he takes on this, where he studies how the capacity of humans to reason can play a central role in achieving this goal of a virtuous and moral life. This fact shows Aristotle as a pioneer of the study of intelligence across the board of human thinking and thus a pioneer of modern AI.

It also shows the importance of looking into Aristotle for possible fresh perspectives since in his work we see how the original ideas on intelligence were developed – tabula rasa – without the possible distortion by the scholarship and science of the centuries that he had brought about!

---

[23] For this second role of Aristotle in AI Ethics, the reader is referred to the recent and white paper [32] of Ober and Tasioulas, "AI Ethics with Aristotle".

## Section 4: Dialectic Rationality and Dialectic Science

We have seen, through the example of ethics, that demonstrative necessity is not appropriate for formulating theories of intelligence that concern multi-criteria problems with context dependent requirements in a real-life setting. Aristotle understood this first. His approach to this class of problems, we have argued, can be very instructive for modern AI, but as we will see in this section this is also true for science more generally.

At the foundational level, the main difference in approaching this type of problems is that we need a more flexible form of reasoning that would allow defeasible reasoning with non-exact[24] knowledge. We need a form of reasoning that would gracefully allow our theory to be non-deterministic where there will be situations in which the theory will be in dilemma between more than one solution, e.g., multiple actions will be equally appropriate but probably for different reasons. This means that we need to move away from absolute rationality and apply a weaker and more flexible form of rationality to govern our theories for this type of problems. Saying this in reverse, multi-criteria problems like that of ethics should not be governed under absolute rationality, as this forces us to select some criteria as strictly dominant, thus going against the natural form of the problem.

In fact, this is what happens in (theoretical) Classical Science. In order to satisfy the demonstrative necessity characteristic of a scientific theory, i.e., that its results will be absolutely certain, the problem undergoes first a Galilean idealization [33] which, amongst other things, means that we have selected some desired subset of dominant criteria, neglecting all others. It is useful to stress this point: there are no real-life laws that are demonstratively absolute, e.g., the laws of Physics can be seen as a good approximation of reality, but not an exact one. They are only exact in a non-real idealized world.

Once we apply this idealization on the problem, we can then find dedicated solutions that maximize the utility for the selected criteria, thus exhibiting optimal rationality, but at any cost, small or large, for the other neglected criteria. In other words, the maxim of science of absolute certainty of results distorts the problem through idealization and simplification, a task that would not be appropriate for other types of problems such as the problems of wider intelligence, e.g., ethics and others in the field of humanities, and indeed for the problem of AI. In fact, it is not clear that this distortion of the problem is appropriate even in the case of hard science, when we consider the pollution effect that results over time due to the negligence of the non-dominant criteria of the problem[25].

---

[24] A way to understand this term of non-exactness, is that the knowledge is highly context dependent and hence it cannot be **compactly** expressed in absolutely exact statements, at least not in the (current) language that we would describe the problem. Nevertheless, these non-exactly expressed statements are useful, as overall they guide us well, provided that we also have a mechanism to guard against the exceptional contexts where they do not apply.

It also helps to remember that scientific theories are human descriptions of reality. Hence there is no a-priori reason why they should be absolute. For example, Quantum Mechanics is such a theory of description at the limit of our knowledge today and hence it is a leap of faith to consider it as absolute. In fact, the uncertainty principle of Quantum Mechanics tells us exactly that they are not absolute – yet the demonstrative science culture still demands that we make it so, at the expense of more and more sophisticated but also perplexing methods even for the specialists in the field [34,35].

How should we then replace absolute rationality? What form should a new rationality take? Drawing from Aristotle's ethics we need a rationality for middle ground and balanced solutions that are sufficiently good and with the potential to sustainably improve. Let us call this **Dialectic Rationality** for obvious reasons, but also because the reasoning form that can serve this form of rationality is that of **Dialectic Argument** of Aristotle. Within Dialectic Rationality the normative condition for a (logically) valid conclusion is that this is **reasonably justified**, i.e., we can build a case for the conclusion that is based on clear facts and assumptions and that the case is internally coherent. In other words, we can bring forward arguments that support the conclusion and that together with additional arguments defend this against possible counterarguments, e.g., arguments for conflicting conclusions.

Aristotle studies this form of argumentative inference extensively with detailed analysis of how we would carry out argumentative debates [36,37]. He studies in detail strategies how to set up a debate and how to win this. But also, the notion of argument is central to his study of reasoning itself – his central notion of a syllogism says explicitly that reasoning is an argument that connects information – and so we can look into his work on Dialectic Argument from a purely reasoning perspective to ask what is a valid logically inference under argumentation. In fact, he states this in the first sentence of the first book of Topics, the collection of books on his works of Dialectic Argument, where he writes:

**_Ἡ μὲν πρόθεσις τῆς πραγματείας μέθοδον εὑρεῖν ἀφ᾽ ἧς δυνησόμεθα συλλογίζεσθαι περὶ παντὸς τοῦ προτεθέντος προβλήματος ἐξ ἐνδόξων, καὶ αὐτοὶ λόγον ὑπέχοντες μηθὲν ἐροῦμεν ὑπεναντίον._**

In free translation, this says:

---

[25] As we have argued here this polluting effect will be even more prominent when we apply theoretical demonstrative science to the humanities, e.g., in social, political, economic science, where there is a growing tendency to do so. Even in philosophical writing, we now see tables with statistics or future projections which are only valid (demonstratively) because they concentrate only on some aspects neglecting all others.

"The purpose of the present treatise is to discover a method by which we shall be able to **reason** from generally accepted opinions about any problem set before us and shall ourselves, **when sustaining an argument, avoid saying anything self-contradictory[26].**"

This is a **universal** statement of "what it is to reason", namely, to build an internally coherent (set of) argument(s) from some premises. Within this view, the internal coherency of arguments becomes the normative condition for validity of any form of reasoning, whether this is in mathematics, science, or reasoning at large. Furthermore, as we have seen above in our brief history of modern AI, the recent study of argumentation in AI resonates closely with this statement of Aristotle, showing argumentation as the wider framework for systematizing human thought, which encompasses, as a limiting case, Aristotle's demonstrative syllogistic logic and the modern Classical Logic of strict logical entailment that emanated from syllogistic logic.

We can therefore move from the dynamically brittle Classical Logic as the underlying calculus of Classical Science to a **Logic of Argumentation**[27], to underly a new form of science, what we have called **Dialectic Science**. Unlike traditional Classical Science, Dialectic Science, under its dialectic rationality, looks for balanced solutions that consider a multitude of criteria. To give an example from Physics, we might be interested how an object falls to stop at the surface of the earth within a reasonable time and a reasonable amount of energy and maybe also with an acceptable amount of heat generation from the friction with the air. This is such a complex problem that even Newton would not have wanted to have to solve! Dialectic Science accepts the difficulty of the problem and aims to find holistic solutions that would at least offer some reasonable guarantees on the various dimensions of interest.

For a classical scientist this dialectic science aim would not look very different, should she had to consider the multi-criterion problem. This is apart from the perspective on these type on holistic solutions. In the Classical Science case this is seen from the perspective of how far we need to move away from perfect and exact solutions to some approximation of these, whereas in Dialectic Science this is seen from the perspective of how far we need to move from necessarily non-perfect solutions to some reasonable level of non-perfection. Indeed, Dialectic Science does not abandon Classical Science – it encompasses it as a limiting case where we can (modulo pollution effects) concentrate on some specific criteria only. For example, in safety critical systems we have a dominant dimension, that of safety, and the solutions to our problem would need to have an absolute guarantee on safety, i.e., the other dimensions need to be plastic so that they can allow safety to remain demonstratively necessary, alas again within an idealized setting that we hope would apply in all real-life cases, but in reality in almost all such cases.

---

[26] The contrary of this statement, namely that when sustaining an argument, i.e., defending against its counterarguments, we say or are forced to accept something contradictory is an invalid argument. This type of self-defeating invalidity is in fact an argumentation-based expression of reductio ad absurdum and the technical link required to show how Argumentation Logic generalizes Classical Logic [38].

[27] We are referring here to reasoning that reaches conclusions from a given knowledge. Other forms of reasoning, that also play an important role in scientific thought, such as inductive and abductive reasoning are meta-forms of reasoning reliant on an underlying form of reasoning to conclusions for their formalization. They are processes that synthesize new knowledge to reason from rather than reasoning processes. Aristotle would consider induction as a reasoning process to a conclusion but we can see that this is a two-step process, first induction generates a generalization as new knowledge and then deductive reasoning with this gives us the conclusion.

So how do we go about Dialectic Science? As hinted above, Aristotle can guide us through his work and understanding of the complex problem of ethics. Let us then go ahead and see AI as an example of Dialectic Science.

**Subsection 4.1: Dialectic Science for AI**

Is Artificial Intelligence a classical science, like others? The historical development of AI together with the momentum of the generally accepted success (modulo its pollution) of science leads us to quickly answer in an affirmative way and even to consider this question as obvious. But intelligence is a multi-facet phenomenon where the standard scientific approach of idealization and simplification cannot be applied. This is unless we do not care at all about its polluting effects, e.g., that we create unnatural or alien forms of intelligence that pollute[28] the human mind rather than enhance it. We cannot ignore the multi-criteria nature of the problem, especially if we want to develop AI systems as thinking machines in a human-centric way that would smoothly integrate in our human cosmos.

The Dialectic Science approach for AI follows the central thesis of Human-centric AI, where the aim is not simply to build artifacts which when used can enhance the capabilities of humans but to build around the human-mind to allow it to evolve in natural and desirable future forms. Unlike traditional Computer Science and Classical Science at large, it takes as its foundational logical basis – its calculus – that of dialectic argumentation [40]. It accepts the necessary multi-dimensional nature of the problem with its difficulty and takes a position away from perfect reasoners (or super-intelligence as it is typically served in the market-driven sector of AI). In fact, Turing, the early pioneer of modern AI, could see this impossibility of intelligent perfection, declaring quite specifically [41] that:

"[…] if a machine is expected to be infallible, it cannot also be intelligent."

Like Aristotle, Turing sees that the problem is too complex to expect to have a demonstrative science for human intelligence.

---

[28] For a discussion of the polluting effects of AI the reader is referred to the article "The pollution of AI" [39].

Let us explore a little the multi-dimensional nature of the requirements for Human-centric AI that makes the problem challenging hard. There are three main high-level categories of quality criteria for the problem of intelligence and its solutions:

**Efficacy**, i.e., a measure of the performance with respect to some utility requirements specified by a particular problem.

**Cognicacy**, i.e., a measure of the understandability and useability of the solution in the human including environment of the problem.

**Ethicacy**, i.e., a measure of the degree of general adherence to human moral values.

Each one of these has its own importance that cannot be ignored as secondary. Classical scientific idealization and simplification cannot be applied. They all need to be considered together in holistic solutions that underly from the very start the development and building of AI. They all need to be part of the design of AI and not an afterthought. Hence the challenge of AI is to find holistic solutions that balance these requirements and that this is done in a sustainable way within the open and dynamic problem environment in which we envisage AI to operate.

This challenge maybe hard, especially when we see it from a traditional classical science perspective where we are deceived to want to find perfect solutions. Within the more realistic approach of Dialectic Science, we are encouraged by the performance of the natural human mind and its intelligence. We can study our intelligence and its role in the human mind's effort to address this challenge - to live a virtuous balanced life, as Aristotle has done, and try to follow its example.

For example, we can recognize that one important characteristic that can support an all-round behaviour of good performance within a good collaborative and ethical life, is that of explainability. Explainability helps in requesting help or collaboration, but also it is a means to be accountable and to solicit useful feedback from which we can learn further about our problem. As such it is recognized as a very important requirement for AI not only because it facilitates the adoption and usefulness of AI, but also because it is essentially necessary for the ethicacy of AI. Indeed, the main requirement on explanations is that these must be **contestable and debatable** [42] in the environment that they operate. Explanations must enable the scrutiny of the reasoning that leads to the proposals they are explaining and indeed the scrutiny of the explanations themselves. Scrutinized by other entities, e.g., by a human who is affected by the decision that is put forward and explained.

Dialectic Science with its foundational basis of dialectic argumentation can address this requirement on explanations in a direct way as **reasoning, argumentation and explanation, essentially coincide**. By directly following Aristotle's universal statement on the validity of reasoning via argumentation, the explanation must make the solution clearly understandable at the cognitive level of the other parties involved so that they can contest the credibility of both the facts and beliefs underlying the solution, as well as the internal consistency of putting these together to support the solution. In this way, an AI explanation must facilitate a **transparent** and **open argumentative debate** on the matter, just like the ones Aristotle studies extensively in the Topics, where the AI system could decide to change its position and produce new solutions with new explanatory justification. Furthermore, the larger aim of these debates would be to facilitate an increasing habitual maturity in the capacity of AI to explain.

Explainability is one example of being informed by the human mind's ability to reason and to cope intelligently in a multi-dimensional and dynamic environment. We have discussed it above because of the close way it relates to dialectic argumentation, the logical foundation of Dialectic Science. But as Aristotle had realized, the most significant lesson we can take from the human mind is to follow its problem-solving strategy of continuous learning on how to balance the many requirements of a "successful life" and to adapt this balance to different circumstances, i.e., that building an all-round intelligence is not a matter of a one-shot process of designing and building an artifact but a habitual process.

The recent developments of AI over that last two decades, namely the development of Machine Learning via Artificial Neural Networks with its prime example of Large Language Models (LLMs), is in phase with this anthropomorphic strategy of addressing the challenge of AI via a continuous process of habitual learning. Unfortunately, this approach is also driven many times by the quest for perfection and super-intelligence, which puts it in the realm of demonstrative classical science[29]. Irrespective of whether such goals could be achieved, they necessarily need the selection of some dominant criterion in order to exhibit their perfection and hence the balance of the solutions over the multitude of criteria will be lost.

Furthermore, it is important to understand that these models of AI, as they are often called, are not theories of intelligence in the sense that they give us some underlying principles under which intelligence operates. For example, LLMs do not form a theory of Natural Language [43,44] but rather they are a good reflection of Natural Language. They are not to be compared for example with Newton's laws of motion but rather with the development of the telescope or microscope. We can build on what they see to apply a theory of intelligence at a practical realistic level, as we saw above in Reasoning about Actions and Change. And yes, they can help us see other things we could not see before, just like a telescope or a microscope, – they are generative. But just like no telescope could have revealed Newton's theory they also cannot reveal by themselves an underlying theory of intelligent reasoning.

---

[29] Furthermore, internally this approach operates under the strict reasoning of maximal probability – a conclusion is drawn only if it can be shown to have maximal probability. This is a very brittle form of demonstrative reasoning depending on the un-informative comparison of numbers where a hair thin difference in the number forces a conclusion. Nevertheless, LLMs are a good reflection of the results of human reasoning in Natural Language.

Nevertheless, this recent approach to AI is in the right direction of Dialectic Science where we can aim to build AI theories of how to learn holistic and adaptable solutions through experience over time. We can adopt a **hybrid approach** of model-centric design with data-centric training. In this some form of explicit design, e.g., ethical **normative guidance**, would be complemented with a process of **training** under data of holistic operation, e.g., data of normative balance between good practical utility and ethically aligned operation. The aim is to train the systems to learn to follow and sustain a holistic approach of seeking, as Aristotle would put it, the middle ground and to maintain through **continuous training** a habitual holistic operation under some normative guidance. Aristotle's notion of leniency guide us to understand that systems should not be designed via strict norms of behaviour that would be absolute but rather that what is needed is flexible normative guidance that helps systems to be ethical in general but also exceptional cases and that learn over time to get better.

In general, Dialectic Science accepts the difficulty of the problem of understanding the many facets of intelligence in some unison form as exhibited in its human existence. It accepts that we may not, as yet, have the appropriate concepts to formulate all round theories of intelligence that would include the element of abstraction and the connection of these theories with our perceived reality of our physical and mental existence. Such concepts and dialectic scientific theories will need to come from a synthesis of the hard sciences and the humanities. They will need the **consilience of human knowledge** [45]. How can we bring together in a consilient way the study of the biology of the brain, the cognitive development of the mind through passive and active training and teaching and the philosophical nurturing of high-level motivations within acceptable limits of morality, to understand how this can give a basis for a smooth and sustainable development of intelligent artificial entities. We need a return to polymathy that would foster a truly inter-disciplinary approach to AI.

**Epilogue**

We have taken a journey into Aristotle's work on human thought from a contemporary perspective of Artificial Intelligence. In this we have tried to understand how his work can help enlighten our thoughts on AI and on Science more generally. The journey has not been one of detail but one that concentrates on the large picture across his work on the essence of human thought and its role in human life. As the original thinker on the systematization of human thought, we have implicitly assumed that by looking at his work we could see afresh the challenge of understanding intelligence. We have then tried to relate this to the today's central challenges of AI.

Aristotle's tour de force study of human reasoning carried the development of science for centuries until our recent times. But when we come to the new challenge of AI, we can see that this trusted path, namely that of Demonstrative Classical Science, is inadequate. It is then useful to go back to the original tabula of Aristotle's work to revisit the central question of what is intelligent reasoning, without any distortion that might have occurred over the centuries.

The dream of reason, as set by Aristotle, was in effect to rationalize reasoning in all its forms as found widely in human thought and expressed through natural language. And with this rationalization of reasoning to rationalize the various endeavors of individual humans and collective society. In some sense this is also the dream of AI today.

We have argued that we can see, within Aristotle's wider work on reasoning and how this relates to human intelligence in the large, a new foundational basis for AI and science more generally. This new foundation is a synthesis of Dialectic Argument as the central notion for rationalizing reasoning – argumentation as reasoning universalis - together with theoretical methodological principles on how to address complex real-life problems, as seen, for example, by a computational view of his work on "practical" ethics. Dialectic argumentation and dialectic rationality serve a new form of Dialectic Science, but one that can be seen as a natural evolution of Classical Science. Dialectic Science is particularly relevant when we are concerned with the pollution effects of science, and therefore we want to address problems in a holistic way that would mitigate the effect of pollution.

At the heart of Dialectic Science stands the need for a **new enlightenment** where we escape from the strong "culture of science" that has developed over the centuries both in the wider scholarly community[30] and the society at large. We need to move away from a culture that considers traditional science as the unquestionable prevailing method of study with the highest possible kudos. This culture of science has become so strong that it has also permeated into many disciplines of humanities – e.g., we have Social Science,

---

[30] This culture of science also permeates in the wider society as a horizontal ideology that covers all other ideologies. It sparked and formed the basis of the renaissance where science will come to help society solve its problems towards a utopian society. But can we claim that it has delivered on this aspiration?

Political Science, Economic Science - pushing studies to include some (usually quite simple, in comparison with those used in the "hard sciences") mathematical or statistical elements so that they can claim a higher kudos based on the demonstrative necessity of their mathematical results[31]. But such absolutely certain results apply to an idealization of the problem, hidden in the many assumptions underlying the mathematical models, rather than its real-life setting, which would become obsolete if some assumption is found wanting.

This strong classical scientific culture of "correct thinking" comes at the price of stifling "free thinking". Instead, in Dialectic Science the emphasis is on accepting the plurality of concerns in real-life problems, like the problem of AI, and the difficulty that comes with this. It aims to address this difficulty via the "free analysis" of the many considerations and parameters that the problem involves. It realistically lowers the expectation of absolute rationality, looking for satisfactory and probably non-perfect solutions in some respects of the problem, but which are sustainably adaptable in a changing dynamic environment. Solutions that will endure in time in contrast of the brittleness of solutions based on demonstrative science of optimality and perfection.

The "beauty of reason" drove the enlightenment eras of the past. This beauty of pure reason has perhaps overestimated the power of reason, blinded by its success when applied in one-sided ways with a primary concern on "profit" in a particular dimension of interest. But maybe in the long-term it is wise to accept a less powerful application of reason in the short-term driven more by long-term benefits for humanity [46].

The most original idea of Aristotle is that of realizing that the power of human reason is in looking for middle ground holistic and habitual solutions to real-life problems in our individual lives and in society at large. The importance of this idea goes beyond the design and development of future AI. It covers future science and more generally how we want to harness the power of thought at in our human existence. It presents us with a new dream of reason away from extremes and quick short-sighted benefits, leading us to a new enlightenment era shaped by humility for the limits of science and human thought and a continuous philosophical reflection on the human cosmos.

**An AI Afterword:** The most enlightening idea of Aristotle particularly for AI, is that of habitual, i.e., continuous long-term, learning and solving of real-life (open) problems. With regards to the question of what is the purpose of AI this idea of Aristotle leads us to look for how we can gradually build AI to help humanity in the long-term. We could work along with the fact of human cognitive fatigue, stemming from the high energy levels needed by the underlying brain processes for cognitive reasoning, and aim to relieve humans from thinking expensively through too much cognition. We could work towards a more efficient, i.e., cheaper in terms of brain processes, form of reasoning by being able to delegate part of, or synthesize, our cognitive reasoning with that of the artificial minds of AI, thus providing real support for the mode of expensive cognitive human thinking.

---

[31] As we have seen above, such studies would cause pollution if their results were adopted except in cases where the study is on a specific aspect of a wider holistic perspective of the problem at hand. This is generally true of Classical Science, i.e., we can use Classical Science for specific components of a problem but do this within the scope of wide and holistic solutions to the problem.

This is what happened with the advent of writing – the first AI technological advance. It helped humans think cheaper in terms of the energy needed for the underlying brain processes and therefore supported the higher-level innovative thinking of humans in science, literature, etc.

**Acknowledgements**

This article came out of the personal desire to be better informed about Aristotle and AI in preparation for the Symposium "Aristotle in the era of AI" held at the Academy of Athens in April 2025. I would therefore like to thank my co-organizers of this event, Kyriakos Demetriou and Doukas Kapantais, for their very valuable help in writing this article. Kyriakos Demetriou has been my philosophy mentor for nearly two decades, without whom this article would never have been written. Doukas Kapantais was always forthcoming with informed answers to my questions on Aristotle. He would point me to the relevant literature thus helping to clarify my understanding and to reduce the time needed to write the article to a reasonable realistic time frame. Many thanks again to both.